\documentclass[letterpaper, 10 pt, conference]{ieeeconf}  % Comment this line out if you need a4paper

\IEEEoverridecommandlockouts                              % This command is only needed if 
\usepackage{booktabs}
\usepackage{tabularx}
\usepackage{graphicx}
\usepackage{subcaption}
\usepackage{balance}
\usepackage{multirow}
\usepackage{makecell}

\usepackage{graphics} % for pdf, bitmapped graphics files
\usepackage{times} % assumes new font selection scheme installed
\usepackage{amsmath} % assumes amsmath package installed
\usepackage{amssymb}  % assumes amsmath package installed
\usepackage{mathtools}
\usepackage{breqn}
\usepackage{float}
\usepackage{hyperref}
\usepackage{comment}
\usepackage{xcolor}

\usepackage{algorithm}
\usepackage{algpseudocode}

\definecolor{comments}{HTML}{ae4dc4}

\newcommand{\name}{GeoSACS}
\newcommand{\longname}{Geometric Shared Autonomy via Canal Surfaces }
\newcommand{\namens}{GeoSACS} %nospace

\title{\LARGE \bf
\name: \longname
}
\author{Shalutha Rajapakshe$^{1,2}$, Atharva Dastenavar$^{1,2}$, Michael Hagenow$^{3}$ \\ Jean-Marc Odobez$^{1,2}$, and Emmanuel Senft$^{1}$ % <-this % stops a space
\thanks{$^{1}$ Idiap Research Institute, Rue Marconi 19, 1920 Martigny, Switzerland
        {\tt\small [srajapakshe|adastenavar|esenft]@idiap.ch}}%
\thanks{$^{2}$ EPFL, Lausanne, Switzerland}
\thanks{$^{3}$ Massachusetts Institute of Technology, Cambridge, Massachusetts, USA
        {\tt\small hagenow@mit.edu}}%
}

\begin{document}

\maketitle
\thispagestyle{empty}
\pagestyle{empty}

%%%%%%%%%%%%%%%%%%%%%%%%%%%%%%%%%%%%%%%%%%%%%%%%%%%%%%%%%%%%%%%%%%%%%%%%%%%%%%%%
\begin{abstract}
We introduce \namens, a geometric framework for shared autonomy (SA). In variable environments, SA methods can be used to combine robotic capabilities with real-time human input in a way that offloads the physical task from the human. To remain intuitive, it can be helpful to simplify requirements for human input (i.e., reduce the dimensionality), which create challenges for to map low-dimensional human inputs to the higher dimensional control space of robots without requiring large amounts of data. We built \name\ on canal surfaces, a geometric framework that represents potential robot trajectories as a canal from as few as two demonstrations. \name\ maps user corrections on the cross-sections of this canal to provide an efficient SA framework. We extend canal surfaces to consider orientation and update the control frames to support intuitive mapping from user input to robot motions.  Finally, we demonstrate \name\ in two preliminary studies, including a complex manipulation task where a robot loads laundry into a washer. 

% Finally, we demonstrate the application of \name\ through two preliminary studies, including complex manipulation tasks such as filling a laundry machine.

\end{abstract}

%%%%%%%%%%%%%%%%%%%%%%%%%%%%%%%%%%%%%%%%%%%%%%%%%%%%%%%%%%%%%%%%%%%%%%%%%%%%%%%%
\section{INTRODUCTION}
%Overall context

Assistive robots provide an opportunity to help users who cannot do specific tasks themselves \cite{brose2010role}. However, building fully autonomous robots for real environments remains a challenge. Frameworks such as shared autonomy (SA) have emerged as a means to blend human inputs with robot control to achieve success in complex tasks \cite{sa_survey,no_to_the_right,coor_shared_autonomy}. SA can allow humans to delegate the bulk of a task to the robot, intervening only when adjustments are necessary for managing uncertainties \cite{corrective_shared_autonomy}. For instance, in a laundry machine loading task, the motion to pick pieces of clothing from a laundry basket and put them in the machine needs to be repeated multiple times. In such a case, the robot could perform the repetitive elements of the task (e.g., moving between the basket and the washer) while the human only provides the required input to grasp the desired clothing article and place it at the right place in the machine. 

%Challenge we want to address
However, common low-dimensional human input interfaces like joysticks can be challenging to map to higher degree-of-freedom (DOF) robot manipulators \cite{6dof_challenges}. 
Although research has explored encoding desired human inputs into a latent space to tackle the dimensionality challenge \cite{latent_actions}, there remain practical considerations for humans to provide this low-dimensional input in SA frameworks. First, such approaches often require gathering large amounts of data (dozens of demonstrations), impractical in real-world settings. Second, the resulting mapping can be hard to comprehend for users \cite{no_to_the_right}.

\begin{figure}[t]
  \centering
  \includegraphics[scale=0.313]{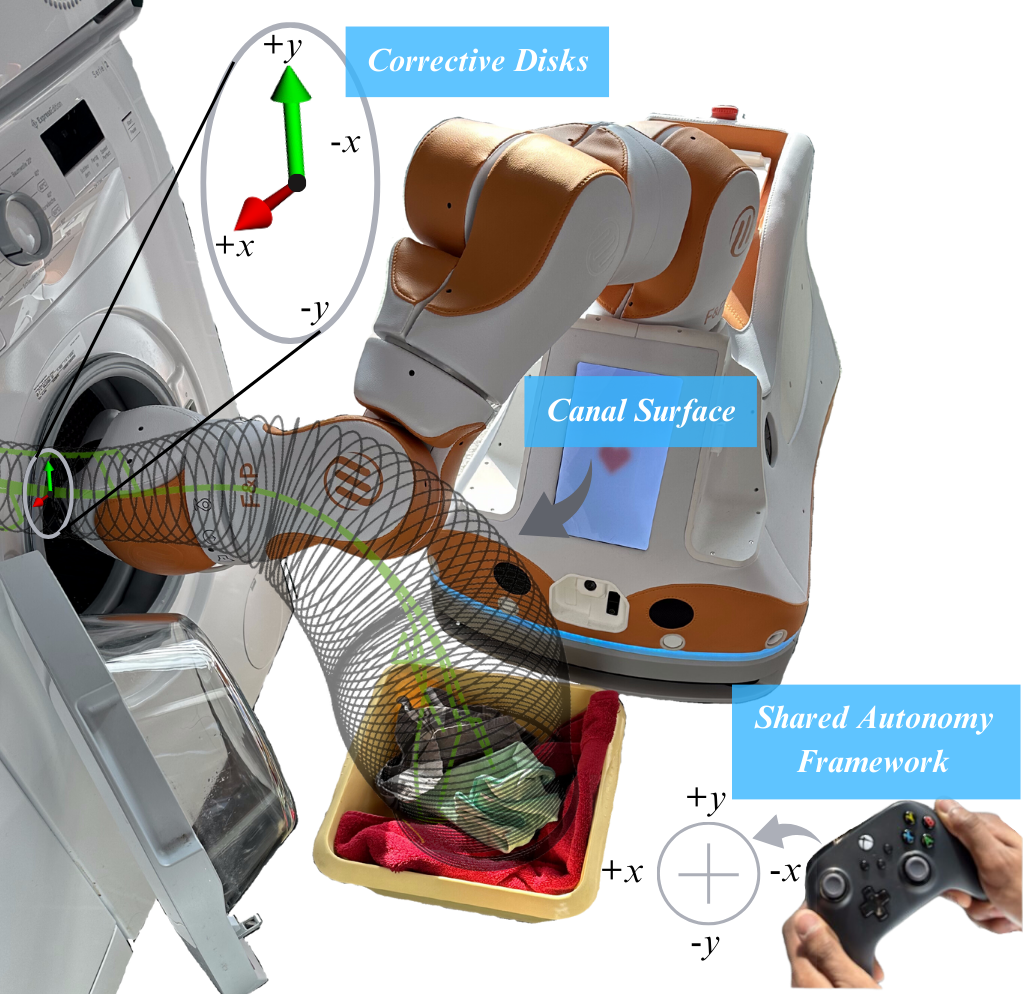}
  \caption{\name\ proposes a SA framework to encode robot motions as canal surfaces while allowing users to provide corrections on the cross-sections of the canal to change the way the robot navigates the canal. Using \namens, users can control robots in real task of daily living, such as filling a laundry machine.}
  \label{fig:teaser}
\vspace{-.3cm}
\end{figure}

%Our method
We introduce \namens, a geometric SA framework with the primary goals of addressing the challenges of data collection and input mapping highlighted previously.
With a minimal number of demonstrations, the task model in \name\ encompasses all the possible variations of the task and lets the user provide corrections to the robot behavior to address task variability. The trajectories are encoded as canal surfaces (see Figure \ref{fig:teaser}) that the robot navigates autonomously, while the user gives corrections on the cross-sections of the canal. The resulting paradigm allows the robot to navigate 6 DOF using a simple joystick and buttons. 
% As the robot navigates the canal, the 2D corrections yield a 3D impact.
%Furthermore, when required, trajectory reversal allows for complex behaviors such as pushing.

%Contributions
In this paper, we present the following contributions:
\begin{itemize}
 
    \item \namens, a SA geometric framework leveraging canal surfaces to map 2D human corrections to the 6D pose of a robot in real time.% This framework aims to ensure continuity and intuitiveness for users, while eliminating the need for large data collection.
        
    \item A practical implementation of the orientation in the canal surface learning framework \cite{trajectory_gen_canal_surfaces} and additional processing to reduce dependency on the quality of the input demonstrations.

    \item A preliminary study demonstrating the feasibility and value of the approach in complex daily tasks such as laundry loading.
    %  for widespread application.

\end{itemize}

\section{Related Work}
Our method builds on canal surfaces and previous SA techniques. To contextualize the contributions of our work, we provide a brief review on (1) learning from demonstration (LfD) and canal surfaces and (2) SA and corrective methods.

\subsection{Learning from Demonstration}\label{related_work1}
LfD is a well-established technique for teaching robots through human demonstrations \cite{lfd_challenges}. Previous work has proposed a range of methods to encode the human knowledge contained in the demonstrations into a robot behavior \cite{osa2018algorithmic}. For example, common approaches include movement primitives (e.g., DMPs \cite{DMPs}, ProMPs \cite{ProMPs}), Gaussian mixture regression \cite{Gmms, improved_GMMs}, dynamical systems (e.g., SEDs \cite{SEDS}), and inverse reinforcement learning \cite{osa2018algorithmic} where a task reward function is inferred from the demonstrations.

One challenge in LfD is how to represent tasks with inherent variability \cite{lfd_survey}. For example, when teaching a robot to load laundry from a basket, the grasp location of the clothing articles in the basket will vary across the pieces, and consequently across the human demonstrations. 
While many approaches consider such uncertainty as part of the LfD task representation (e.g., probabilistic methods or object-centric methods), ideally an LfD method intended to be part of a human-in-the-loop system would model the envelope of uncertainty (i.e., the range of actions a person might want to perform).
A promising LfD approach \cite{trajectory_gen_canal_surfaces,trajectory_gen_GC} that naturally meets these requirements employs canal surfaces \cite{canal_surfaces}, capturing the range of data present in a set of demonstrations. This method combines a curve (i.e., mean behavior) with a tangent surface that bounds the variability observed in the demonstrations, with a focus on teaching robots tasks such as pick-and-place and obstacle avoidance.

\subsection{Shared Autonomy}\label{related_work2}
In SA, a human and robot work together to complete shared goals \cite{sa_survey}. Some of the common SA approaches include providing assistance based on human goal inference, dynamically allocating roles (i.e., control) between a human and robot, and providing informed corrections to robot behaviors as they execute tasks \cite{corrective_shared_autonomy}. 

Most relevant to our work are SA methods relying on user corrections that have gained popularity recently \cite{no_to_the_right, corrective_shared_autonomy}. Corrections have been used in SA both as a way to learn over an uncertain robot reward function and to facilitate effective human real-time interaction with robots.
% with an otherwise autonomous robot behavior. 
%\cite{corrective_shared_autonomy}. 
For example, Cui et al. \cite{no_to_the_right}, correct a robot's action space based on a combination of voice commands and joystick-guided actions. 

% Similarly, Ahmadzadeh et al. \cite{trajectory_gen_GC} present a method for physical corrections in robots using canal surfaces.

% Similarly, Ahmadzadeh et al. \cite{trajectory_gen_GC} present a method for physical corrections in robots using canal surfaces, however, may not be suitable for those with limited mobility. 
% In contrast, our approach emphasizes remote corrections to enhance accessibility.

For SA systems leveraging remote or latent space input, one important design decision is the input mapping, or how user input maps to corrections in the robot's space. Previous work in teleoperation has demonstrated how a poor input mapping can negatively impact user performance \cite{coor_frames, consistency_control_frames}. Ideally, this mapping would be intuitive, consistent, and minimize any mental transformations required by the operator \cite{mental_transformations}. Existing methods in SA have addressed the challenges of input mapping in a limited way (e.g., through heuristics or optimization objectives in learning control mappings \cite{corrective_shared_autonomy,contr_robot_latent_actions}), however, more general solutions remain an open challenge. We believe that canal surfaces, where user corrections can occur on the cross-sections of the canal, present an opportunity to ground the operator input mapping during SA.

\section{Background on Canal Surfaces} \label{sec:previous_work}

\subsubsection{Definition}\label{sec:canal_surfaces}

\begin{figure}[h]
\centering
\begin{subfigure}{.15\textwidth}
    \centering
    \includegraphics[width=.95\linewidth]{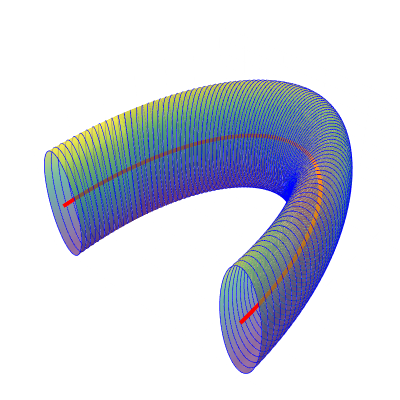}  
    \caption{}
    \label{canal1}
\end{subfigure}
\begin{subfigure}{.15\textwidth}
    \centering
    \includegraphics[width=.9\linewidth]{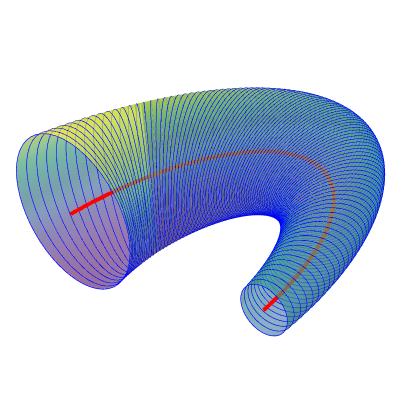}  
    \caption{}
    \label{canal2}
\end{subfigure}
\begin{subfigure}{.15\textwidth}
    \centering
    \includegraphics[width=.95\linewidth]{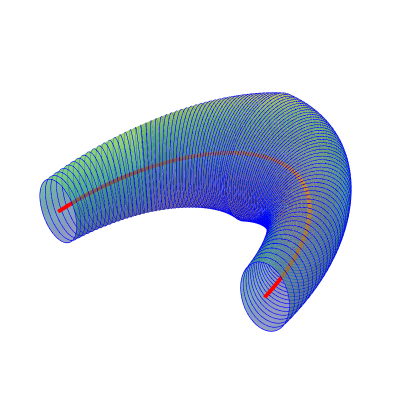}  
    \caption{}
    \label{canal3}
\end{subfigure}

\caption{Circular cross-section canal surfaces with the same directrix curve and different radii functions. The cross-section is fixed to a circular shape, and the radii can change along the directix as shown in (b) and (c).}
\label{fig:canal_surfaces}
\end{figure}

The concept of canal surfaces introduced by Hilbert and Cohn-Vossen \cite{canal_surfaces} presents a geometric way of representing surfaces and their properties. When a sphere of varying radius moves along a path, the resulting surfaces of the spheres, traces out the canal surface. In the context of robotic control, previous work \cite{trajectory_gen_canal_surfaces, trajectory_gen_GC} directly use a discretized version of canal surfaces where the canal is represented as a series of circular cross-sections (referred to as ``disks'') orthogonal to the tangent vector of a regular curve $\Gamma$ : \( \textbf{x} \) = \( c(s) \)  $\in$ $\mathbb{R}^{3}$ (called as the \textit{directrix}) in 3D Cartesian space, with $s$ denoting the discrete state value. The \textit{radii} function \( r(s) \) $\in$ $\mathbb{R}$ denotes the radius of the disk, $C_s$ at $s^{th}$ point on the directrix, see Fig. \ref{fig:canal_surfaces} for three example of canal surfaces.

% \begin{figure}[tbph]
%   \centering
% %   \framebox{\parbox{3in}{We suggest that you use a text box to insert a graphic (which is ideally a 300 dpi TIFF or EPS file, with all fonts embedded) because, in an document, this method is somewhat more stable than directly inserting a picture.
% % }}
%   \includegraphics[scale=0.25]{figures/canal_surface_sample.png}
%   \caption{Sample canal surfaces with varying sphere radii}
%   \label{fig:canal_surfaces}
% \end{figure}

% A canal surface is generated by moving a sphere of varying radius along a space curve, called the spine curve, where the radius of the sphere can change as it moves. The surface is the envelope of this family of spheres. The concept finds application in various fields, such as computer graphics, animation, and the design of complex shapes in engineering.

% The basic idea is that at each point along the spine curve, a sphere is placed whose radius can vary from point to point. The canal surface is then the set of all points that lie on at least one of these spheres. When the radius of the sphere is constant, the canal surface becomes a special case known as a tube or swept sphere surface. Canal surfaces can represent a wide variety of shapes and are particularly useful for smoothly blending shapes or creating organic forms that are difficult to model with more conventional geometric primitives.

\subsubsection{TNB frames}\label{sec:tnb_frames}

Frenet-Serret, or TNB (for Tangent, Normal, Bi-normal), are coordinate systems typically used to describe the geometry at a given point on a curve in 3D space and consist of 3 mutually orthogonal unit vectors: \textbf{e}$_{T}$ (toward the curve direction), \textbf{e}$_{N}$ (toward the curve center), and \textbf{e}$_{B}$ (binormal vector orthogonal to both $\textbf{e}_{T}$ and $\textbf{e}_{N}$). In the context of canal surfaces, the plane defined by  $\textbf{e}_{N}$ and $\textbf{e}_{B}$, which is orthogonal to $\textbf{e}_{T}$, serves as the basis for representing the disks.

However, TNB frames are prone to noise with real data, and in instances where the \textbf{e}$_{N}$ vanish can happen making it unusable. Addressing this, previous work on canal surfaces for LfD \cite{trajectory_gen_GC} has adopted an improved technique known as Bishop frames~\cite{bishop}, which employs the concept of relatively parallel fields. %Unlike TNB frames, Bishop frames do not require the computation of torsion, making them more practical. 
Bishop frames retain the tangent vector concept from TNB frames but calculate the normal (\textbf{e}$_{N}$) and binormal (\textbf{e}$_{B}$) vectors to minimize rotational changes, ensuring smoother transitions. This process involves selecting an initial vector at the start of the curve and then rolling it along the curve without slipping or twisting. As a result, this approach tends to maintain the orientation of the frame relative to the curve's geometry, providing a smooth transition along the curve.

In the context of SA, Bishop frames offer smoother transitions compared to TNB frames, however, the frequent directional changes observed in Bishop frames along the curve can complicate the intuitive understanding of this mapping, as illustrated in Fig. \ref{fig:correction_axes}. %This observation highlights that Bishop frames alone would not be sufficient as a user's input at one point may produce an outcome significantly different at neighbouring points on the curve. 
Therefore, in \S \ref{sec:shared_autonomy_framework:input_mapping}, we introduce a novel control framework aimed at providing an intuitive mapping of user inputs by aligning when possible the normal vector \textbf{e}$_{N}$ with a global x-axis.

\subsubsection{ Canal Surface Navigation} \label{traj_repro}

% In order to reproduce a trajectory, the robot will start from its usual starting position such that the initial position $p_{0}$ of the end effector lies on the first circular cross sectional plane $\mathbf{C}_{0}$. Then following the method outlined in \cite{} we measure the distance $\overline{p_{0}\textbf{c}_{0}}$ and divide it from $r(0)$ which is the radius of the corresponding circular cross section in order to obtain the ratio $\eta = \overline{p_{0}\textbf{c}_{0}}/\overline{g_{0}\textbf{c}_{0}}$. This $\eta$ is used to generate the next pose of the end effector on the next circular cross section using the equation,

% According to the ``ratio rule'' described in \cite{trajectory_gen_canal_surfaces}, 

When navigating the canal surface (as $s$ being incremented), previous work use the ``ratio rule'' \cite{trajectory_gen_canal_surfaces} to accommodate the varying nature of radii along the curve. 
%To navigate along the disks of the canal surface, a method that accommodates the varying nature of radii along the curve is required. The ``ratio rule'' \cite{trajectory_gen_canal_surfaces}, aims for accomplishing this. 
It maintains a constant ratio of the distance from a point on a circle to its center, relative to the circle's radius. Through finding this ratio for a given disk, we can transfer the point onto the next, preserving the ratio despite variations in the radii. In \S \ref{sec:shared_autonomy_reproduction}, we leverage this rule within our proposed framework to reproduce trajectories.

\section{Methodology}

The key insight of this paper is that canal surfaces are a powerful framework for SA. We propose \namens, a SA framework where canal surfaces are learned from a minimum of two demonstrations, then the robot autonomously navigates the resulting canal surface while receiving human corrections. We map 2D corrective user inputs onto the disks composing the canal surface, influencing the robot movements in the 3D space. Although we consider the orientation values captured through demonstrations in our framework, we do not define the effect of corrections on orientation. This section presents our modifications to the canal surface framework presented in \cite{canal_surfaces} to support SA.

% that we do not impose any restrictions on the orientation during the demonstration phase. However, the effect on the robot’s orientation is not explicitly define

% \ES{list here the assumptions: no need to change orientation along circular cross sections, any orientation between the demonstration is valid}

%The corrections will not influence the orientation, however, the robot will adhere to the orientation information captured during the demonstrations as discussed in section \ref{sec:orientation_integration}.

\subsection{Integrating Orientation Data into Canal Surfaces} \label{sec:orientation_integration}

Although, canal surfaces have been utilized to represent robotic trajectories within LfD frameworks \cite{trajectory_gen_canal_surfaces, trajectory_gen_GC}, they only encapsulate positional data and omit orientation. %Therefore, we introduce a method to consider orientation data within these canal surfaces. 
% Canal surfaces solely capture positional data, inherently limiting trajectory reproduction to these constraints. 
While orientation can be critical to complete a task, requiring the end effector to attain a defined orientation can lead to issues such as limited reachability or position tracking \cite{relaxedIK}. 
As such, to integrate orientation in our canal surface framework, for each disk, we use the demonstrations to compute the mean orientation which represent the target orientation for a disk and standard deviation (SD) which represents the permissible range of orientation variation for a given disk. Details on finding these values are discussed in \S \ref{sec:mean_orientation}.

\subsection{Input Mapping} \label{sec:shared_autonomy_framework:input_mapping}

To map our 2D corrections in the canal surface, we applied the human input provided at each state onto the disk representing the space within the canal surface. 
However, TNB or Bishop frames' $\textbf{e}_{N}$ and  $\textbf{e}_{B}$ are under-constrained and subject to large changes over time, as such there are concerns to directly employ them as correction axes. Instead, we define a new correction frame that prioritize the consistency of user inputs between frames.
% Employing TNB or Bishop frames' $\textbf{e}_{N}$ and  $\textbf{e}_{B}$  as correction axes observed to be non-intuitive. Consequently, there is a pressing need to establish two axes on the disks of a canal surface in a more consistent way, ensuring users can conveniently identify the impact of their inputs on the robot's movements.
Similarly to TNB and Bishop frames, our first axis is the tangent vector $\textbf{e}_{T}$. The remaining two axes are positioned on the plane orthogonal to $\textbf{e}_{T}$. Of these axes, we define one as the ``correction x-axis'' $\textbf{x}_{s}$, which is aligned as closely as possible with the fixed global x-axis $\textbf{x}_{G}$, while ensuring smooth continuity. The other axis, the ``correction y-axis'' $\textbf{y}_{s}$, is orthogonal to $\textbf{x}_{s}$ and secures local alignment by maintaining continuity with the y-axes of previous disks. 

For convenience, we adopt the notation $\textbf{e}_{T}(s)$ to represent the tangent vector at the $s^{th}$ point on the directrix and $C_s$ to denote the disk orthogonal to $\textbf{e}_{T}(s)$ at that same point. Here, $s$ ranges from $1$ to $N_f$, with $N_f$ indicating the total number of points on the directrix.

\subsubsection{Correction x-axis on \(C_{s}\)}

% \subsubsection{Correction \normalfont x\textit{-axis on \(C_{s}\)}} 

% For the \(s^{th}\) circular cross section \(C_{s}\), to determine the correction x-axis \( \mathbf{x_{s}} \) with a smooth transition from the correction x-axis \( \mathbf{x_{s-1}} \)  on the \((s-1)^{th}\) cross section \(C_{s-1}\) , we follow a 3-step process.

To determine the correction x-axis \( \textbf{x}_{s} \) on \(C_{s}\), with a smooth transition from the previous correction x-axis \( \textbf{x}_{s-1} \), while maintaining global interpretabiliy, we blend the projection of \( \textbf{x}_{G} \) and \( \textbf{x}_{s-1} \).

First, we find \( \textbf{x}^{C_{s}}_{G} \), the projection of \( \textbf{x}_{G} \) \textit{on} \(C_{s}\)
    % Given that we are at the \(s^{th}\) circular cross section \(C_{s}\), our first step involves extracting the tangent vector \( \mathbf{\textbf{e}}_{T}(s) \) which is orthonormal to \(C_{s}\) .
using the Equation \ref{eq:projection_equation}, which serves as an initial estimation for \(\textbf{x}_{s} \).

\begin{equation}
    \textbf{x}_G^{C_s} = \textbf{x}_G - \left( \textbf{x}_G \cdot \textbf{e}_{T}(s) \right) \textbf{e}_{T}(s),
    \label{eq:projection_equation}
\end{equation}

Then, we find \normalfont \( \textbf{x}_{s-1}^{C_{s}} \), the projection of \( \textbf{x}_{s-1}\) \textit{on} \(C_{s}\) 
%Depending on the orientation of the disks, relying solely on \( \textbf{x}_{G}^{C_{s}} \) to define \( \textbf{x}_{s} \) may result in abrupt directional shifts. This lack of continuity in \( \textbf{x}_{G}^{C_{s}} \) along the trajectory, renders it inconsistent and non-intuitive for user corrections. To mitigate this effect, we consider the preceding correction x-axis, \( \textbf{x}_{s-1} \) in the computation of \( \textbf{x}_{s} \). Hence, we project \( \textbf{x}_{s-1} \) onto \(C_{s}\), in order to obtain \( \textbf{x}_{s-1}^{C_s} \) by 
using the same equation as before.%\ref{eq:projection_equation}. 

Finally, we use spherical linear interpolation (Slerp) from  \( \textbf{x}_{G}^{C_{s}} \) to \( \textbf{x}_{s-1}^{C_{s}} \) to generate \( \textbf{x}_{s} \) while balancing smooth transition and alignment with \( \textbf{x}_{G}\).
%We aim to prioritise \( \textbf{x}_{G}^{C_{s}} \) and \( \textbf{x}_{s-1}^{C_{s}} \) based on the orientation of \(C_{s}\) relative to the   \( \textbf{x}_{G}\) as aligning  \( \textbf{x}_{s} \) with $\textbf{x}_G$ is our primary goal while keeping the continuity. 
%To achieve this, we use Equation \ref{weighting_formula}, which effectively tries to position $\textbf{x}_{s}$ aligning with $\textbf{x}_{G}$, however keeping the continuity. We use a circular interpolation between the vectors (also called slerp) with a ratio based on the angle between the xG and XCG.
Our slerping ratio (interpolation constant) is determined based on the angle between the $\textbf{x}_{G}^{C_s}$ and $\textbf{x}_{G}$ as it can inform whether $\textbf{x}_{G}^{C_s}$ actually contains useful information (e.g., if $\textbf{x}_{G}^{C_s}$ and $\textbf{x}_{G}$ are orthogonal, then $\textbf{x}_{G}^{C_s}$ direction is mostly arbitrary and as such could be mostly disregarded). However, we would like to emphasise that other heuristics can be used to set the interpolation constant. 

% However, we would like to emphasise that other interpolation metrics (such as dot product) could be used here.

% \begin{equation}
% % \textbf{x}_{s} = \frac{\sin((1 - t) \cdot  \theta) \cdot \textbf{x}_{G}^{C_s} + \sin(t \cdot  \theta) \cdot \textbf{x}_{s-1}^{C_s}  }{\sin(\theta)},
% \textbf{x}_{s} =
% \begin{cases}
%     \frac{\sin((1 - t) \cdot  \theta) \cdot \textbf{x}_{G}^{C_s} + \sin(t \cdot  \theta) \cdot \textbf{x}_{s-1}^{C_s}  }{\sin(\theta)}, & \text{if } \theta \neq 0 \, \text{and} \, \theta \neq \pi \\
%     \textbf{x}_{s-1} & \text{else } 
% \end{cases}
% \label{weighting_formula}
% \end{equation}

\begin{equation}
% \textbf{x}_{s} = \frac{\sin((1 - t) \cdot  \theta) \cdot \textbf{x}_{G}^{C_s} + \sin(t \cdot  \theta) \cdot \textbf{x}_{s-1}^{C_s}  }{\sin(\theta)},
\textbf{x}_{s} =
\begin{cases}
    \frac{\sin((1 - t) \cdot  \theta) \cdot \textbf{x}_{G}^{C_s} + \sin(t \cdot  \theta) \cdot \textbf{x}_{s-1}^{C_s}  }{\sin(\theta)}, & \text{if } \text{sin}(\theta) > \epsilon \\
    \textbf{x}_{s-1} & \text{else } 
\end{cases}
\label{weighting_formula}
\end{equation}

where $\theta = \cos^{-1}\left(\frac{\textbf{x}_{s-1}^{C_s}}{\Vert \textbf{x}_{s-1}^{C_s} \Vert} \cdot \frac{\textbf{x}_{G}^{C_s}}{\Vert \textbf{x}_{G}^{C_s} \Vert} \right)$ denotes the angular distance between the projected vectors, $t = \frac{\theta_X}{\pi/2}$ is the interpolation constant, and $\theta_X = cos^{-1} \left( \frac{\textbf{x}_{G}^{C_s}}{\Vert \textbf{x}_{G}^{C_s} \Vert} \cdot \textbf{x}_G \right)$ denotes the  alignment of $C_s$ with $\textbf{x}_G$.

% $\theta_X$ is a measure of the alignment of $C_s$ with $\textbf{x}_G$. 

As $\textbf{x}_{G}^{C_s}$ is the projection of $\textbf{x}_G$, $\theta_X$ will be confined to have a maximum value of $\frac{\pi}{2}$ rad.
% The $\theta_X$ obtained by taking the arccosine between $\textbf{x}_{G}^{C_s}$ and $\textbf{x}_{G}$ is a measure of the alignment of $C_s$ with $\textbf{x}_G$ and falls within the range of $[0, \frac{\pi}{2}]$. 
A smaller \( \theta_X \) measures significant alignment of $C_s$ with \( \textbf{x}_{G} \), which in turn should prioritize more on $\textbf{x}_{G}^{C_s}$ over continuity with \( \textbf{x}_{{s-1}}^{C_{s}} \). Nevertheless, to prevent sudden shifts from \( \textbf{x}_{s-1} \) to \( \textbf{x}_{s} \), the Equation \ref{weighting_formula} effectively limits the shift even if it means slightly compromising on the ideal alignment with \( \textbf{x}_{G} \). 
Conversely, a larger \( \theta_X \) suggests a lower alignment of $C_s$ with \( \textbf{x}_{G} \). This indicates a greater emphasis on the continuity of \( \textbf{x}_{s} \) with \( \textbf{x}_{s-1} \) compared to aligning  \( \textbf{x}_{s} \) with \( \textbf{x}_{G} \).

\subsubsection{Correction y-axis on \(C_{s}\)} 

% \subsubsection{Correction \normalfont y-axis \textit{on} \(C_{s}\)} 

The correction y-axis $\textbf{y}_{s}$ on $C_s$ is determined based on local alignment to ensure smooth continuity. In a first step, we generate a preliminary $\textbf{y}^{'}_{s}$ by computing the cross product of $\textbf{e}_T(s)$ and $\textbf{x}_{s}$. This result would be correct and sufficient most of the time. However, this reliance of cross product can create issues. For example, when going up after picking an object, $\textbf{e}_{T}(s)$ would flip, while $\textbf{x}_{s}$ maintains the same direction. This partial flip would result in $\textbf{y}_{s}$ flipping too which could be highly unintuitive for users. This reversal would most of the time occurs over few steps. 
%Using the previous correction y-axis $y_{s-1}$ to align continuity can be an option. Nonetheless, there are common exceptional cases to consider. For example. ideally, $\mathbf{e}_T(s)$ should directly flip to align with sudden directional changes, especially in dynamic scenarios such as end-effector going down and suddenly up in a picking motion, going left and then suddenly right, as in inserting a cloth into laundry machine. However, the imperfections in real-world demonstrations can cause a gradual directional shift of the tangent vectors across closely spaced disks to reach the final opposite directional tangent vector, rather than an instant flip. Relying solely on $y_{s-1}$ is ineffective in such scenarios as the gradual transition of tangent vectors does not adequately capture the directional shifts. While these transitions appear gradual across closely spaced disks, they ultimately constitute an abrupt directional change when viewed collectively over this short range. 
Therefore, to promote stability and consistently follow the preceding trend, we employ a window-based strategy, averaging the correction y-axes of the previous 10 disks and project the averaged (mean) vector onto the current disk $C_{s}$ to derive $\textbf{y}_{\mu}^{C_{s}}$ using the Equation \ref{eq:projection_equation}. 

The angle $\theta_{Y}$ between $\textbf{y}_{\mu}^{C_{s}}$ and $\textbf{y}^{'}_{s}$, on $C_{s}$, is then measured. If $\theta_{Y}$ exceeds $\frac{\pi}{2}$ radians, suggesting a major deviation and discontinuity, we invert $\textbf{y}^{'}_{s}$ to realign with the historical y-axes trend. If the angle is less than $\frac{\pi}{2}$ radians, $\textbf{y}^{'}_{s}$ is considered sufficiently aligned and remains unchanged.
% The window size of 10 was empirically found to strike an optimal balance between responsiveness and stability in our canal surface model. However, 
For trajectories with denser directrix points, a larger window size may be advantageous to further smooth out orientation shifts.

\begin{figure}[h]
  \centering
  \includegraphics[scale=0.16]{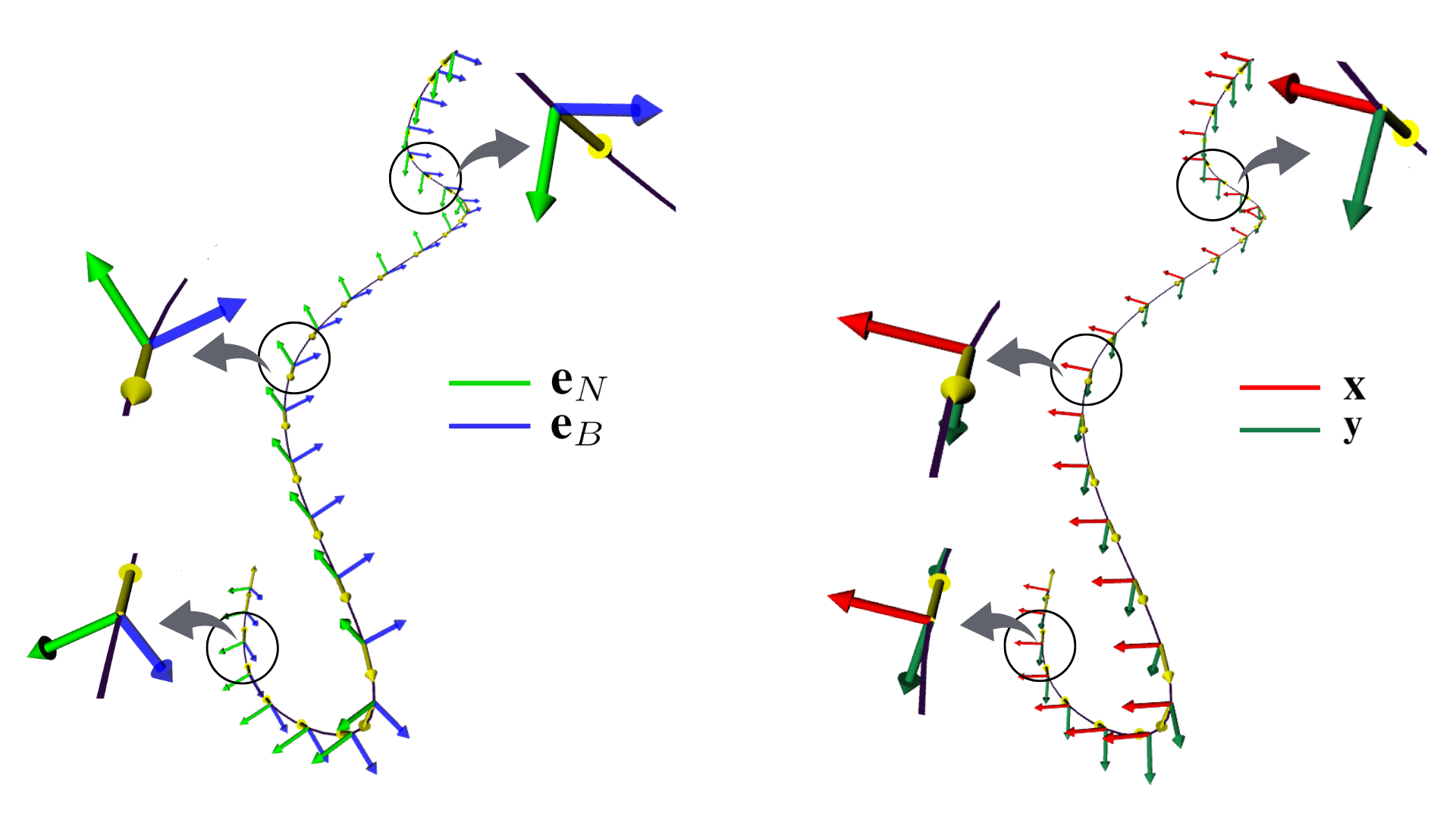}
  \caption{Frame evolution of Bishop frames (left) and our correction frames (right) along a curve with tangent vector illustrated in yellow. Highlighted points illustrate how our correction frames exhibit increased consistency and reduced directional changes compared to Bishop frames.}
  \label{fig:correction_axes}
  \vspace{-.3cm}
\end{figure}

With this process, we define the correction frame on \(C_{s}\) as the combination of \( \textbf{e}_T(s) \), \(\textbf{x}_s\), and \(\textbf{y}_s\). Fig. \ref{fig:correction_axes} displays the correction frames generated along a directrix curve.

% To define \( \textbf{y}_{s} \), the correction y-axis on \(C_{s}\), we create a first estimate \( \textbf{y}^{'}_{s} \) by taking the cross product between  \( \textbf{e}_T(s) \) and \( \textbf{x}_{s} \), resulting in a vector lies on \(C_{s}\). However, it is important to note that when deriving \( \textbf{y}_{s} \) solely from \( \textbf{e}_T(s) \), abrupt changes in direction (e.g. end effector moving down then suddenly up, as in a picking motion) can disrupt \( \textbf{y}_{s} \)'s continuity. To mitigate this, we employ a window-based approach, averaging the 10 previous correction y-axes (Line 11 in Algorithm \ref{algo}) and projecting this average \( \textbf{y}_{m}\) onto \(C_{s}\) as per the Equation \ref{eq:projection_equation} to find \(\textbf{y}_{m}^{C_{s}}\) (Line 12 in Algorithm \ref{algo}). We then measure the angle \(\theta_{Y}\) between \(\textbf{y}_{m}^{C_{s}}\) and \( \textbf{y}^{'}_{s} \) in which both are on $C_s$ (Line 13 in Algorithm \ref{algo}). If this angle exceeds \( \frac{\pi}{2}\)rad, indicating a significant deviation and lack of continuity, we flip \( \textbf{y}^{'}_{s} \) to better align with the preceding correction y-axes. Conversely, if the angle is under  \( \frac{\pi}{2}\)rad, \( \textbf{y}^{'}_{s} \) is deemed sufficiently aligned and remains unchanged (Line 14 in Algorithm \ref{algo}). With this, we define the correction frame on \(C_{s}\) as the combination of \( \textbf{e}_T(s) \), \(\textbf{x}_s\), and \(\textbf{y}_s\). Fig. \ref{corraxes} displays the correction frames generated along a directrix curve.

% \begin{figure}[tbph]
% \centering
% \begin{subfigure}{.22\textwidth}
%     \centering
%     \includegraphics[width=\linewidth]{figures/tnb_final.png}  
%     \caption{Bishop frames along a directrix curve}
%     \label{tnbframes}
% \end{subfigure}
% \begin{subfigure}{.22\textwidth}
%     \centering
%     \includegraphics[width=\linewidth]{figures/corr_final.png}  
%     \caption{Correction frames along a directrix curve}
%     \label{corraxes}
% \end{subfigure}

% \caption{A comparison illustration of the proposed mapping framework with Bishop frames}
% \label{fig:correction_axes}
% \end{figure}

% \begin{figure}[tbph]
%   \centering

%   \includegraphics[scale=0.25]{figures/transition.png}
%   \caption{Generated correction axes for different tasks }
%   \label{fig:correction_axes}
% \end{figure}

\subsection{Trajectory Reproduction} \label{sec:shared_autonomy_reproduction}

\begin{figure}[t]
  \centering
  \includegraphics[scale=0.28]{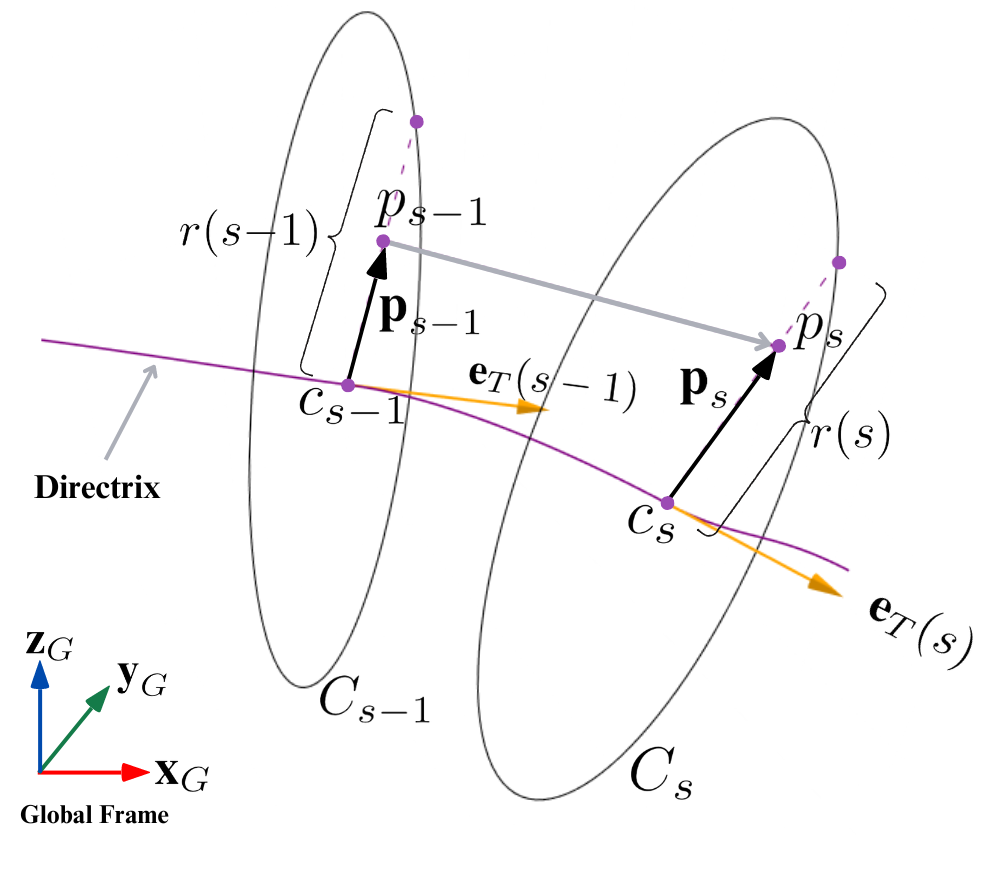}
  \caption{ Visual representation of the trajectory reproduction phase from frame $s{-}1$ to frame $s$. 
}
  \label{fig:reproduction}
\end{figure}

After constructing the canal surface with integrated orientation data and correction frames, users can now leverage the complete pipeline to execute tasks. Trajectory generation adheres to the ratio rule, as elaborated in section \ref{traj_repro}. The robot starts from the position \(p_{s-1}\) pointed by the vector $\textbf{p}_{s-1}$ situated on \(C_{s-1}\) (refer to Fig. \ref{fig:reproduction}). Then the distance \(\overline{c_{s-1}p_{s-1}}\) is measured and divided by  \(r(s{-}1)\), representing the radius of \(C_{s-1}\). This division results in the ratio \(\eta = \overline{p_{s-1}c_{s-1}}/ r(s{-}1)\) which is used to determine the next position vector $\textbf{p}_{s}$, which points to $p_{s}$ on the subsequent disk \(C_{s}\) using the following equation:

\begin{equation}
    \textbf{p}_{s}  = \eta \cdot r(s) \cdot \left( \textbf{R}_{s}^{s-1} \cdot \frac{\textbf{p}_{s-1}}{\Vert \textbf{p}_{s-1} \Vert}\right),% + (\textbf{c}_{s} - \textbf{c}_{s-1}),
\end{equation} 

where $\textbf{R}_{s}^{s-1}$ represent the rotation between $C_{s-1}$ and $C_s$.
% and $\textbf{c}_{s}$ and $\textbf{c}_{s-1}$ denotes the translation vectors pointing to $c_s$ and $c_{s-1}$ on the directrix.
    
%where $'$ refers to the homogeneous coordinate (e.g., $\textbf{x}' = [\textbf{x}^T \quad 1]^T$)
% where $H^{s}_{s-1}$  denotes the homogeneous transformation matrix allowing to obtain point \(p_s\) from point $p_{s-1}$.
%and $H^{s}_{s-1}$  the homogeneous transformation matrix transforming $\textbf{p'}_{s-1}$ into $C_s$ in order to obtain \(\textbf{p}_s\). %Additionally, for multiplication purposes, we augment the vectors with a homogeneous coordinate,
%\( \textbf{p'}_{s-1} = [\textbf{p}_{s-1} \quad 1]^{T} \) to obtain \( \textbf{p}_{s} = [\textbf{p}_{s} \quad 1]^{T} \) and then extracts the vector part for the next steps.

% $\textbf{p}_{s-1} = [\textbf{p}_{s-1} 1]^T $ to obtain $\textbf{p}_{s} = [\textbf{p}_{s} 1]^T $, and then extracts the vector part for the next steps.

% $p_s$ between the correction frames on \( C_{s-1}\) and  \( C_{s}\), and \(r(s)\) denotes the radius of \(C_{s}\).

% between the frame formed by $\textbf{e}_{T}(s-1), \textbf{x}_{s-1}, \textbf{y}_{s-1}$ on \( C_{s-1}\) and  the frame formed by $\textbf{e}_{T}(s), \textbf{x}_{s}, \textbf{y}_{s}$ on \( C_{s}\), and \(r(s)\) denotes the radius of \(C_{s}\).

Instead of maintaining a fixed ratio value across all disks, an adaptive ratio strategy can be employed based on specific requirements. As outlined in \cite{trajectory_gen_GC}, an exponential decay function can be utilized for this purpose:

\begin{equation}
    \eta_s =  (\eta_{0} - \eta_{f})e^{-\lambda (s-1)} + \eta_{f}
\end{equation}

where ${\eta}_{s}$ represents the ratio of $C_s$, and $\lambda$ denotes the decay constant. Utilizing this equation, the ratio can smoothly transition from its initial value $\eta_0$ to the desired value $\eta_f$. For instance, when $\eta_f = 0$, the reproduced trajectory gradually converges towards the directrix. 

% Conversely, setting $\eta_f = 1$ leads to a divergent strategy, where the trajectory aims to deviate and follow a path closer to the boundary of the canal.

% where i+1 represents the ratio at step ui , and γ denotes
% the decay constant. Utilizing this equation, the ratio can
% smoothly transition from its initial value η0 to the desired
% value ηf . Moreover, when ηf is set to zero, the reproduced
% trajectory adopts a convergent strategy, gradually aligning
% with the path towards the directrix. Conversely, setting ηf =
% 1 leads to a divergent strategy, where the trajectory aims to
% deviate and follow a path closer to the boundary of the canal

\subsection{Integrating Corrections} 

\begin{figure}[h]
  \centering

  \includegraphics[scale=0.25]{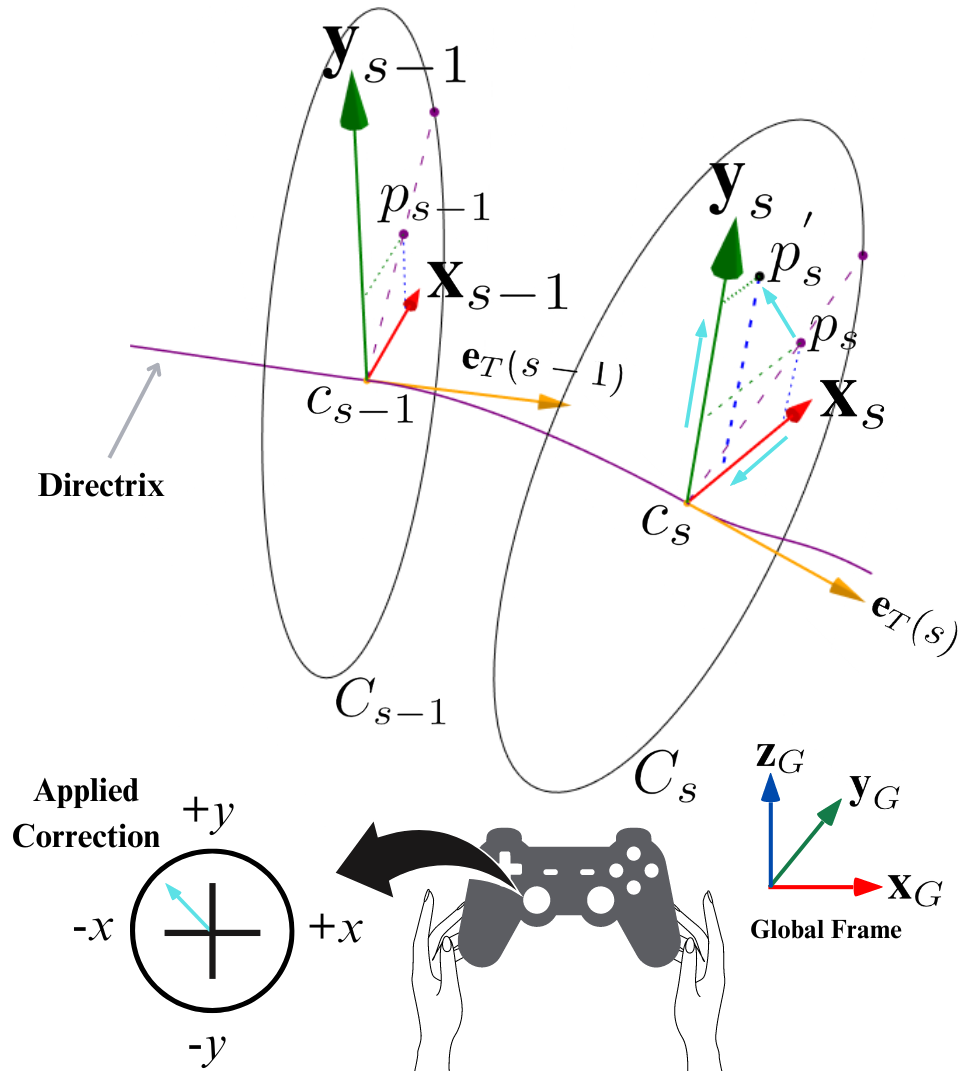}
  \caption{Applying a correction on \(C_{s}\) (blue arrow marked on the joystick control pad).  The x and y axes marked on the joystick control pad control the corrections on  \(C_{s}\) along \( \textbf{x}_{s} \) and \( \textbf{y}_{s} \) respectively. The blue arrows marked on \(C_{s}\) indicate how the input of the joystick affects \(p_{s}\) to reach \(p^{'}_{s}\).}
  \label{fig:correction_apply}
\end{figure}

During runtime, users have the flexibility to give corrections at any moment via a joystick. If a user intervenes when the robot end-effector is at $p_{s}$, subsequent robot motions during that correction period are determined by the joystick input and are restricted to $C_s$. Details of joystick input integration can be found in \S\ref{sec:joystick_integration} and Fig. \ref{fig:correction_apply} illustrates how corrective user inputs influence $p_s$ on $C_s$ to reach \(p^{'}_{s}\). Once the correction period ends, a new trajectory is computed from \(p^{'}_{s}\) with the same method as outlined in \S\ref{sec:shared_autonomy_reproduction} and restrictions on motion are lifted to move along the subsequent disks. 

\subsection{Backtracking for Repetitive Tasks} \label{sec:shared_autonomy_backtracking}

Finally, to complete repetitive tasks, when a motion has to be repeated multiple times, we support trajectory reproduction in the opposite direction of the demonstrated data. For example, for a pick-and-place task, the backtracking would allow the robot to return to the starting position and restart the process.
Finally, we also allow the user to trigger this backtracking to more easily recover from unsuccessful motion and redo part of the motion. Complemented with the 2D correction planes, this feature allows users to make corrections in 3D space, providing comprehensive control over the robot's movements. To achieve this, we reverse the sequence of disks in the canal surface model, whilst maintaining the same correction frames. Trajectory reproduction is then achieved through the same way as previously discussed. We leverage this feature in real-time, creating a loop between the demonstrated pick and place locations, allowing the user to switch direction of motion, and provide corrections in both directions with the same mapping.

\section{Implementation}

We implemented \name\ on a robotic system containing Lio robot \cite{lio_overview} by F\&P Robotics, an assistive robot designed for eldercare or home environments, and a joystick to input corrections.  The empirically determined values for smooth control during implementation are listed in Table \ref{tab:constant}. 
Our code can be found online \footnote{ \url{https://github.com/ShaluthaRajapakshe/geosacs.git}}.

\subsection{Preprocessing Demonstrations}
\subsubsection{Recording}
With the formulation explained in \S \ref{sec:previous_work}, generation of canal surfaces can be done using multiple demonstrated trajectories. For our approach, we used kinesthetic teaching. During the demonstrations, we record the 3D Cartesian position and the orientation of the robot's end effector at 20Hz. For the $m^{th}$ demonstrated trajectory $T^{m}$, comprising $N_m$ data points, each single data point is formatted as $\hat{\phi^m} = \left(x^m, y^m, z^m, q^m_w, q^m_i, q^m_j, q^m_k\right)$, where the position components $(q^m_x, q^m_y, q^m_z)$ belong to $\mathbb{R}^3$, representing Euclidean space, and the orientation components $(q^m_w, q^m_i, q^m_j, q^m_k)$ are expressed as a quaternion. This structure applies to all demonstrations, indexed from $m = 1$ to $n$. As our system targets repetitive pick and place tasks, demonstrations start from the pickup location, and end at the placing location.

\subsubsection{Preprocessing Position Data}
For each demonstration, we cannot assume that $N_m$ will be consistent. Previous work on LfD involving canal surfaces has utilized interpolation and resampling, while suggesting dynamic time warping (DTW) \cite{dtw} as an alternative method to align demonstrations. However, our findings indicate that employing either method in isolation does not suffice to produce a canal surface with the desired smoothness. Therefore, we incorporated a two-step filtering method. 

% Intially, we apply DTW for temporal alignment of trajectories, then filter based on the warping path indices, yielding filtered trajectories $T^{m}_{f1}$ with an equal count of points $N_{f}$. Next, we perform a second filtering phase to reduce noise from sudden overshoots in trajectories, caused due to human motor control limitations in kinesthetic demonstrations. We do this by applying a cubic spline to sample data points via a step filter (refer to Equation \ref{step filter}) of step size \(h = \alpha \cdot N_f\), where \(\alpha = 0.1\) (empirically set and can be adjusted based on the requirement). 

First, we apply DTW for temporal alignment of trajectories. Next, we perform a second filtering phase to reduce noise from sudden overshoots in trajectories, caused due to human motor control limitations in kinesthetic demonstrations. This involves the use of a cubic spline to sample data points through a step filter.  The step size $h$ is empirically determined to be 10\% of the total number of points on the trajectory. Using this step size, we then sample points along the trajectory. Finally, a cubic spline is fitted over the sampled points and a smoothed trajectory with $N_{f}$ points is produced through resampling.

% \begin{equation}
%     T^{m}_{f2} = \{ T^m_{f1}(i) \,|\, i \, \% \, c = 0, \, \forall i\in \mathbb{N} \cap [0,N_f-1] \}
%     \label{step filter}
% % \end{equation}

% where \( T^{m}_{f2}\) denotes the two step filtered final trajectory points for the $m^{th}$ demonstration. 

% Finally, a cubic spline is fitted over \(T^{m}_{f2}\), and a smoothed trajectory with $N_{f}$ points is produced through resampling.

\subsubsection{Preprocessing Orientation Data}

For orientations, a similar methodology is followed. After applying the DTW algorithm for positional data, we extract quaternion-based orientation values corresponding to the same filtered trajectory points. Using the same step size $h$, we then sample orientation values accordingly. To interpolate these points, we utilize Catmull-Rom quaternion interpolation,\footnote{Splines library was employed for Catmul-Rom quaternion interpolation: \url{https://pypi.org/project/splines/}} a method balancing linear interpolation's simplicity and Slerp's complexity, to ensure smooth transitions. Finally, we resample at the same points used for positional data to maintain consistency in the data set. The final result is a structured dataset \(T_f\) of $n$ demonstrations, each with \(N_f\) data points, represented as $\phi^{7 \times N_f \times n }$ including both position and orientation data.

\subsection{Canal Surface Generation}

After obtaining the pre-processed data, the next step is to generate the canal surface.

\subsubsection{Directrix} To find the directrix $\Gamma$, where the centers of all disks lie, we calculate the mean value $\mu_p$ across the positional coordinates of \(T_f\), where $\mu_p \in \mathbb{R}^{3 \times N_f}$.

\subsubsection{Radii} Secondly, the boundaries of the disks that form the canal surface along the directrix are determined by calculating the radii of these disks, which indicates the task's spatial limitations. For instance, trajectories converging in a confined space suggest restricted movement within that zone. %Therefore, by identifying the radii, we can define the boundaries and the spatial constraints within the operational environment.

Following previous work \cite{trajectory_gen_canal_surfaces}, we calculate the radius at a specific directrix point by measuring the distances between this point and the corresponding points on the filtered trajectories in  \(T_f\). We then designate the largest of these distances as the radius for the focal point. This approach guarantees that all subsequent points fall within the boundary set by the circle with this radius, thus defining the permissible action space at that location.  After determining the directrix and the radii for the disks, we then identify the correction frames for each point on the directrix, as detailed in Section \ref{sec:shared_autonomy_framework:input_mapping}.

\subsubsection{Orientation Measures}\label{sec:mean_orientation}

We obtain the mean quaternion $\bar{\textbf{q}}_s$ and associated SD \(\sigma_{q_{s}}\) for \(C_{s}\) by leveraging the processed demonstration data in \(T_f\). 
% following the methodology explained in \S \ref{sec:orientation_integration}. 
We use an Eigen value-based method  to determine the mean \cite{sckitLib}, and a measure of absolute distance to ascertain the SD\footnote{Pyquaternion library was utilized to find the SD of orientation within a disk: \url{https://kieranwynn.github.io/pyquaternion/}} across the demonstrations. This SD is crucial for dynamically adjusting the cost of orientation constraints in our optimization-based inverse kinematics (IK) algorithm, ensuring the end effector can achieve specific points within the orientation constraints. Details on this approach are discussed in \S \ref{sec:motion_gen}.

% \begin{equation}\label{eq:mean_orientation}
%      A = \frac{1}{n} \sum_{i=1}^{n} q_i \otimes q_i 
% \end{equation}

\subsubsection{Cross Section Refinements}

In pick and place tasks, it may occur that parts of the disks lie in collision areas, especially when the picking (or placing) action is done on a horizontal surface (common in human environments). Due to the kinesthetic nature of demonstrations,  the canal surface geometry in these areas is likely to not feature tangent vectors perfectly orthogonal to the horizontal surface, hence leading to parts of the disks being below the surface that the robot might try to reach. Therefore, we identify the disks where this situation occurs and correct the local canal surface geometry for safe robot  operation. If \(C_s\) features this issue, we compute the angle between  \( \textbf{e}_T(s) \) and the global z-axis, \( \textbf{z}_G \). Depending on the measured angle, we assign \( \textbf{e}_T(s) \) to \( \textbf{z}_G \) or \( -\textbf{z}_G \). Finally, we appropriately adjust the corresponding correction frame.

% project \( \textbf{e}_N(s) \) onto the horizontal plane  and compute the new \( \textbf{e}_B(s) \) from the cross product of two previously computed vectors.

\begin{table}[h]
  \caption{Constant values used in our implementation. }
  \label{tab:constant}
  \vspace{-0.05in}
  \setlength\tabcolsep{4pt}
  \begin{center}
  \begin{tabular}{l|lllllll}
    \toprule
    
    \textit{Parameter} & $\epsilon$ & $\lambda$ & $w_{\text{p}}$ & $\alpha$ & $\beta$ & $b$ & $\delta$ \\
    \textit{Value} & 1e-10 & 5e-4 & 100 & 9 & 0.3 & 15 & 150\\

\bottomrule
  \end{tabular}
  \vspace{-0.3cm}
\end{center}
\end{table}

\subsection{Motion Generation} \label{sec:motion_gen}

Our optimization-based IK engine, inspired from RelaxedIK \cite{relaxedIK}, seeks a joint configuration $\mathbf{\Theta}$ that balances accurate end effector position ($\textbf{p}_s$) and orientation ($\bar{\textbf{q}}_s$) by minimizing the cost function $J(s)$:

\begin{equation}
J(s) = w_{\text{p}} \cdot J_{\text{p}}(\textbf{p}_s, \mathbf{\Theta}) + w_{\text{o}}(s) \cdot J_{\text{o}}(\bar{\textbf{q}}_s, \mathbf{\Theta}),
\end{equation}

where $J_{\text{p}}$ and $J_{\text{o}}$ are the cost functions for positional and orientational errors, and $w_{\text{p}}$ and $w_{\text{o}}(s)$ are the weighting factors for each cost component.
% (empirically set values can be found in Table \ref{tab:constant}). 
% emphasizing the relative importance of position and orientation accuracy on \(C_{s}\).

For $w_{\text{o}}(s)$, we assign a higher weight for a lower SD in the demonstrations, indicating a narrow allowable range of orientation variability, and a lower weight  for a higher SD, suggesting less stringent constraints on orientation, using the following equation:

\begin{equation}
w_{\text{o}}(s) = e^{-\alpha(\sigma_{q_s}-\beta)},
\end{equation}

where $\alpha$ and $\beta$ are empirically determined to ensure that $w_{\text{o}}(s)$ falls within a specific range $[0, b]$ to keep the balance between position and orientation while $w_{\text{o}}(s)$ changes. %In our implementation, by setting $\alpha = 9$ and $\beta = 0.3$, we restrict the maximum  $w_{\text{o}}(s)$ to be $b = 15$ when $\sigma_{q_s}=0$.

\subsection{Joystick Integration}\label{sec:joystick_integration}

Sampled at 20hz, joystick inputs are provided as scalar values $k_x$, $k_y$ within the range [-1,1]. We obtain the resulting corrective displacement \( \textbf{d}_s\) on $C_s$ with the following equation:

\begin{equation}
\textbf{d}_s =  \frac{k_x}{\delta}\textbf{x}_s + \frac{k_y}{\delta}\textbf{y}_s
\end{equation}

where $\delta$ is a scaling factor influencing the sensitivity, and $\textbf{x}_s$, $\textbf{y}_s$ are the correction axes on $C_s$. As long as user intervention lasts, $\textbf{d}_s$ is recomputed at the control frequency, added to the current point in the disk as illustrated in Fig. \ref{fig:correction_apply}, and the resulting position is sent to the robot's IK engine. 

% \ES{add a small section about IK, control, and correction - how do we go from joystick value between 0-1 to a correction in the disk between 0-1 }

\subsection{Canal Parametrization}

% We define a new parameter $d$ whose discrete values vary from -1 to 1 and include 0. These respectively designate the first, last, and middle disks of the generated canal. From how we provide our demonstrations (from the picking to the placing location), these respectively correspond to three defined locations : \textit{pick}, \textit{place} and \textit{home}.

To efficiently navigate within the canal surface, considering the previously introduced backtracking, we define a discrete parameter \(d\) that ranges from -1 to 1, where -1 representing the start of the motion (the \textit{pick} action), 0 the middle of the motion (the \textit{home} position), and 1 the end of the motion (the \textit{place} action).%, and middle (\textit{home}) disks of the canal, aligned with demonstration stages from picking to placing locations.

Along with the backtracking feature explained in \S \ref{sec:shared_autonomy_backtracking}, we leverage this parameter to create a loop between \textit{pick} and \textit{place} locations with different ratio strategies during the trajectory reproduction as discussed in \S \ref{sec:shared_autonomy_reproduction}. The robot cycles continually through these locations in the following order: \textit{home - pick - home - place - home}. In order to enhance user interaction, any trajectory heading towards \textit{home} will adopt a convergent ratio strategy, whereas trajectories towards \textit{pick} and \textit{place} will adopt a fixed ratio strategy. This ratio strategy resets the corrections between the pick and place actions without preventing users to provide corrections during the pick and place actions.

\begin{figure*}[t]
    \centering
    \begin{subfigure}[b]{0.48\textwidth}
        \centering
        \includegraphics[scale=0.18]{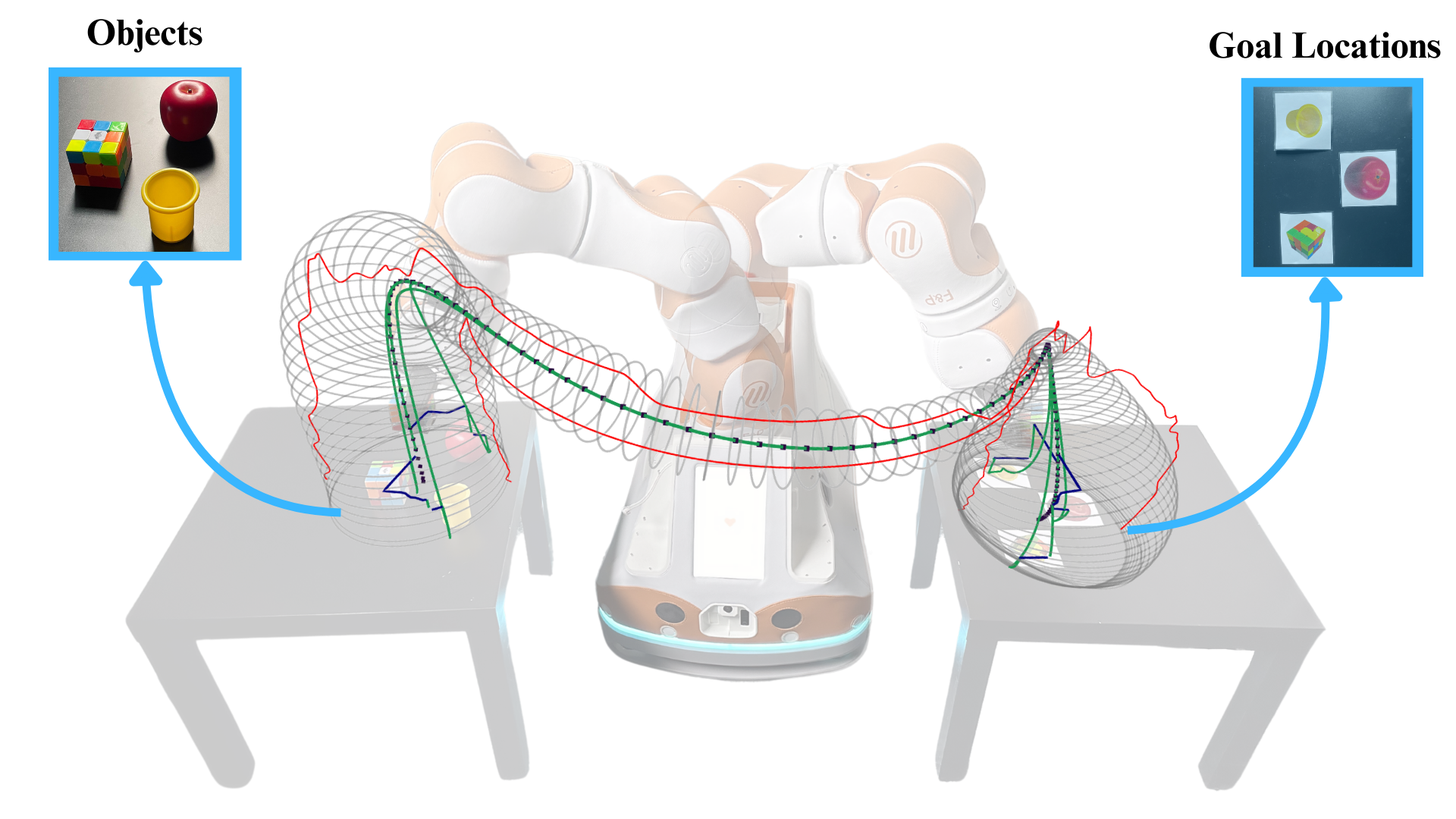}
        \caption{Task 1 workspace and trajectories}
        \label{fig:experiment1}
    \end{subfigure}%
    \hfill
    \begin{subfigure}[b]{0.48\textwidth}
        \centering
        \includegraphics[scale=0.18]{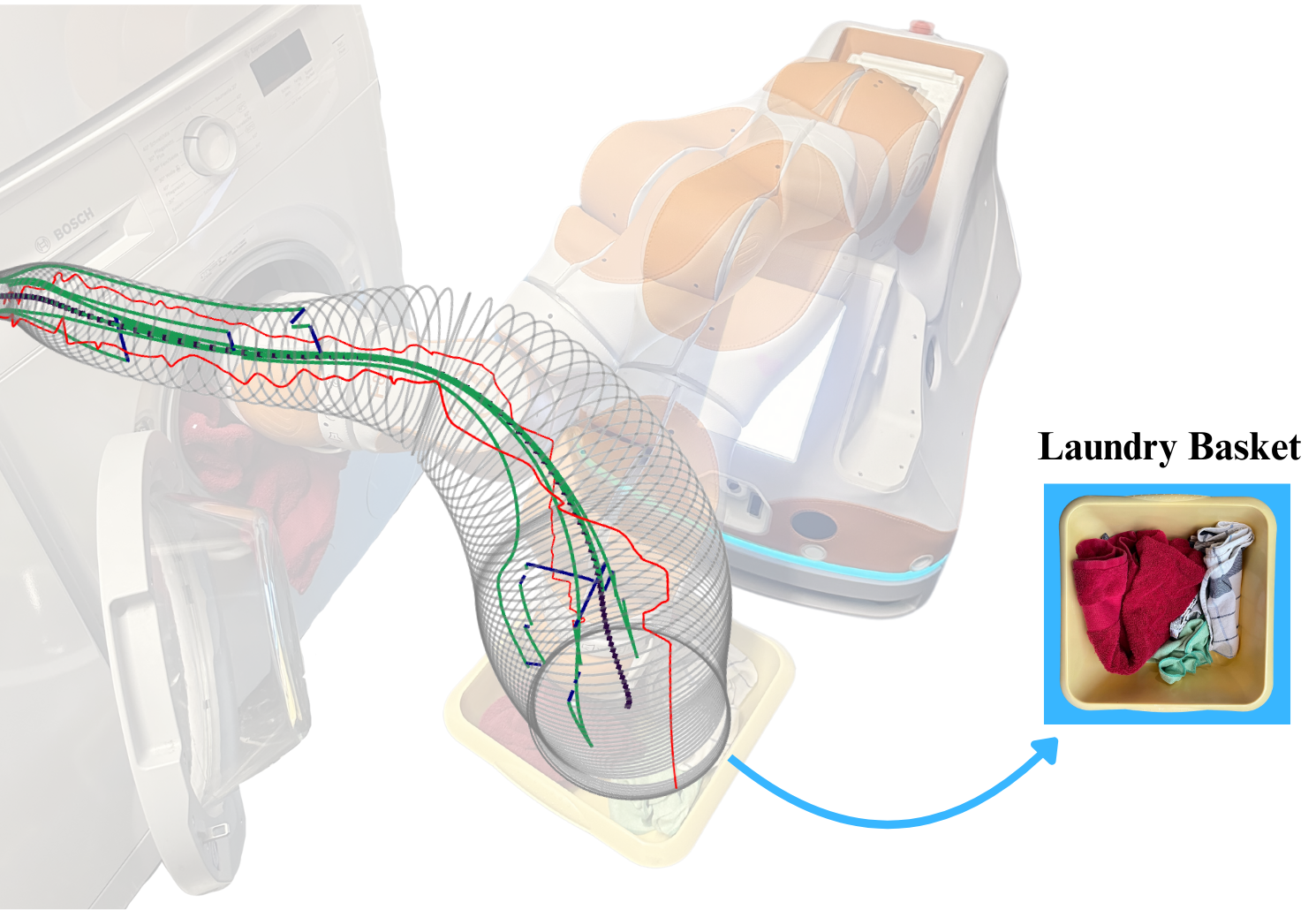}
         \caption{Task 2 workspace and trajectories}
  \label{fig:experiment2}
    \end{subfigure}

    \caption{Experiment work-spaces and the observed correction trajectories; The red curves show the original demonstrations, dotted blue curve shows the directrix, black circles represent the disks, green curves show the generated trajectories, and the blue solid lines indicate the corrections.}
    \label{fig:experiments}
\end{figure*}

\section{Preliminary study}

To demonstrate \namens, we designed a preliminary study showcasing two tasks inspired by household activities. The first is a pick-and-place task between two tables, requiring objects from one table to be placed at specific locations on the other. The second task involves realistic laundry loading. Each task was performed by two expert operators (part of the authors). In the future, we plan to conduct user studies to achieve a more comprehensive and nuanced evaluation.

\subsection{Tasks Description}

\subsubsection{Task 1: Targeted Object Relocation}

The initial task entails picking up and positioning three items; an apple, a cup, and a cube into specific target spots. Illustrated in Fig. \ref{fig:experiment1} including the generated canal surface from two demonstrations, these objects are initially situated on one table, with their designated destinations displayed on a second table. Throughout the trials, we altered the starting points and the goal locations of the objects.

% \begin{figure}[tbph]
%   \centering
%   \includegraphics[scale=0.204]{figures/Experiment1_workspace.png}
%   \caption{Task 1 workspace: red curves show original trajectories, blue curve shows the directrix, and black circles show canal surface cross sections.}
%   \label{fig:experiment1}
% \end{figure}

\subsubsection{Task 2: Laundry Loading}

The second task aims to showcase the practical application of our method in a routine activity, here loading a washing machine. Users are required to transfer three differently sized clothing items from the basket to the machine in any order. Fig. \ref{fig:experiment2} illustrates the task setup, emphasizing the narrow confines of the workspace and the canal surface generated after two demonstrations.

% \begin{figure}[tbph]
%   \centering
%   \includegraphics[scale=0.24]{figures/laundrySample.png}
%   \caption{Task 2 workspace: red curves show original trajectories, blue curve shows the directrix, and black circles show canal surface cross sections.}
%   \label{fig:experiment2}
% \end{figure}

\subsection{Results}
One user provided demonstrations for both tasks, and both users  completed successfully the two tasks. The supplemental videos\footnote{Recordings of all experiments including the demonstrations procedure is available at: \url{https://www.youtube.com/playlist?list=PLd5mNkwFUMTiRjwFC3P53dVPu709jcY_s}} showcase of how the users performed the two tasks including the demonstration procedure. Fig. \ref{fig:experiments} illustrates the evolution of the trajectories following corrections by an expert operator for the two tasks.

\subsubsection{Task 1: Targeted Object Relocation}

For task 1, the average completion time (across two trials with two users) was observed to be 2 minutes and 55 seconds, with a SD of 5.5 seconds. On average, users spent 16.6\% of the total task time providing corrections. 
% However, task 1 sometimes saw minor deviations when positioning objects within the markers.
% \ES{Any interesting observation? - hard to see objects with gripper}

We noted a challenge for the operators, particularly  with the gripper's side view, where only one of its two fingers was visible, obscuring the gripper's full extent. This made it difficult for operators to accurately judge the depth for grasping objects. Notably, the users did not need to use the backtracking feature during this task.

\subsubsection{Task 2: Laundry Loading}

For task 2, the average completion time was 2 minute and 18 seconds, with a SD of 2.5 seconds, and an average of 13.6\% of the task time was spent on corrections. 

Users employed the backtracking feature to push clothes further inside the laundry drum that were partially hanging out of the laundry door after the initial attempt, showcasing a practical real-life scenario. Adding orientation data to canal surfaces proved beneficial for inserting clothes into the laundry machine, as without orientation guidance, the robot tended to collide with the machine.

\section{Limitations and Future Work}

%\subsection{Observations}
We observed from our preliminary study that \name\ allowed our operators to successfully control the robot while requiring only a limited number of demonstrations and corrections. Anecdotally, both users required a short training to understand the mapping before being proficient. 
% \subsection{Limitations and Future Work}
Despite the successful demonstration of feasibility, our approach and the preliminary study have some limitations.

\subsubsection{Method Limitations}

\paragraph{Cross section refinements}
Although our refinements prevent collisions with horizontal surfaces, these modifications are task specific and can introduce undesirable discontinuities in disk orientation. We aim to find a generalizable approach that can also be applied to vertical surfaces and provide smooth transitions with the rest of the canal.

\paragraph{Correction Y axes}
We prioritize local alignment of the correction y-axes to maintain consistency and continuity. However, axis shifts may still arise if there's a gradual change in axes orientation. For instance, in task 1, despite the correction x-axes consistently pointing in the same direction, the correction y-axes shifted at the picking and placing locations due to a gradual direction change of the tangent, which pointed up during picking and down during placing.

\paragraph{Orientation integration}
Joystick corrections directly influence the position and indirectly affect orientation via our IK engine based on each disk's SD. Future work will focus on improving orientation control for users. (e.g., through a second joystick, voice, or other modalities).

% \ES{you can add here the other points we mentioned}
%orientation could be different (e.g., interpolation)

\subsubsection{Study Limitations}
The main limitation of our study was the use of a limited number of expert users on fixed tasks. With this preliminary study, we aimed to demonstrate the feasibility of our approach for real-world complex tasks such as laundry loading. In the future, we plan to conduct a larger scale user study involving participants with disabilities to assess the efficiency and intuitiveness of our approach.

Secondly, although it is envisioned as a means to support users with disability, we did not include such users in the initial development of the method. In the future, we plan to conduct participatory design sessions including users with disabilities to ensure that this system meets their specific needs in terms of expressivity and interaction modalities.

Thirdly, we also plan to explore how such a SA framework can be used for remote teleoperation. As the view will be provided from a camera integrated in the gripper, this extension will require significant changes in the control frames that we will be exploring in future work.

%Within a SA framework, the proposed geometric method successfully integrated corrections to complete the two tasks. With the addition of mean orientation supported with the proposed pre-processing steps, we were able to generate promising canal surfaces even with two demonstrations. Given these preliminary results, we believe our contributions can be utilized for performing daily activities with the support of a human.
% Furthermore, we plan to conduct a more thorough user study and a participatory design phase to identify and improve our method. Finally, we aim to investigate how our method can be adapted to people with reduced mobility and their caregivers and in the context of remote teleoperation.

\section{Conclusion}

In this paper, we presented a SA framework leveraging canal surfaces to encode variability within repetitive behavior. We demonstrated the application of our method in two preliminary studies where expert users controlled the robot to complete home tasks such as filling a laundry machine. Furthermore, we showed that only two demonstrations were sufficient to capture the main behavior components and the variability required to complete the tasks.

% \begin{table}[h]
% \caption{An Example of a Table}
% \label{table_example}
% \begin{center}
% \begin{tabular}{|c||c|}
% \hline
% One & Two\\
% \hline
% Three & Four\\
% \hline
% \end{tabular}
% \end{center}
% \end{table}

%    \begin{figure}[thpb]
%       \centering
%       \framebox{\parbox{3in}{We suggest that you use a text box to insert a graphic (which is ideally a 300 dpi TIFF or EPS file, with all fonts embedded) because, in an document, this method is somewhat more stable than directly inserting a picture.
% }}
%       %\includegraphics[scale=1.0]{figurefile}
%       \caption{Inductance of oscillation winding on amorphous
%        magnetic core versus DC bias magnetic field}
%       \label{figurelabel}
%    \end{figure}



\section*{ACKNOWLEDGEMENT}

This research has received funding from the Loterie Romande as part of the CollabCloud project.

%%%%%%%%%%%%%%%%%%%%%%%%%%%%%%%%%%%%%%%%%%%%%%%%%%%%%%%%%%%%%%%%%%%%%%%%%%%%%%%%

\bibliographystyle{IEEEtran} % We choose the &quot;plain&quot; reference style
{\footnotesize
\bibliography{refs} % Entries are in the &quot;refs.bib&quot;
}

\end{document}